# MoEXViz: Interactive Visualizations of Sparse Mixture-of-Experts

Chinmayee Raman, Hazim Hasan Bajamal, Reza Rawassizadeh
Department of Computer Science,
Metropolitan College, Boston University

## Abstract

Sparse Mixture-of-Experts architectures have become a prominent strategy for scaling large language models without proportionally increasing inference cost. By routing each token through a sparse, input-dependent subset of expert subnetworks, these models increase total parameter capacity while limiting active computation. However, their fragmented routing pathways make them difficult to inspect, audit, and debug.

We present MoEXViz, an open-source, browser-based visualization platform for analyzing sparse MoE models at the per-token and per-layer levels. MoEXViz separates model execution from visualization through a three-stage pipeline: offline PyTorch extraction, compact JSON serialization, and low-latency browser playback. This design removes the need for live GPU access during interactive analysis while preserving traceability to real model forward passes. MoEXViz supports multiple visualization views, including expert UMAP projections, per-layer expert activation grids, average activation trends, per-domain expert rankings, prompt-level inspection, and animated routing diagrams. The system dynamically derives layouts from model metadata rather than relying on hard-coded expert counts or routing widths. We demonstrate MoEXViz on three architecturally distinct open-weight MoE models: OLMoE-1B-7B, JetMoE-8B, and DeepSeek-MoE-16B.

Code is available at https://github.com/Resource-Efficient-AI-Lab/MoEXViz and a live demo is available at https://resource-efficient-ai-lab.github.io/MoEXViz/.

## 1 Introduction

Transformer-based large language models (LLMs) [1] have achieved strong performance across a wide range of natural language processing tasks. However, as these models scale to hundreds of billions of parameters, dense Transformer architectures face substantial computational, memory, and energy costs during both training and inference.

Sparse Mixture-of-Experts (MoE) architectures [2], [3] address this challenge by replacing dense feed-forward network layers with multiple expert subnetworks and a learned router. For each token, the router selects only a small subset of experts, allowing the model to increase total parameter capacity while keeping active computation relatively low. This sparse, input-dependent routing makes MoE models efficient, but it also makes their internal behavior harder to understand.

Recent research demonstrates that standard FFN layers function as key-value memories. In particular, the first projection acts as "keys" that detect syntactic and semantic patterns, while the second projection acts as "values" that map those patterns back to vocabulary predictions [4], [5]. Because only a small subset of memory slots activates for any given token [6], sparse MoE models replace these uniform lookups with dynamically sharded memory banks. In the context of MoE, the routing network functions as an active retrieval query, dynamically selecting a sparse top-k set of experts based on input embeddings [7], [8].

This transition mirrors biological modularity, where localized processing structures evolve to maximize task fitness while minimizing wiring costs [9]. Similarly, MoE gating networks, when optimized under load-balancing and capacity constraints [2], [9], can drive experts to specialize during training. These experts may differentiate into distinct semantic, morphological, or syntactic niches without requiring the full model capacity for every token [4], [10].

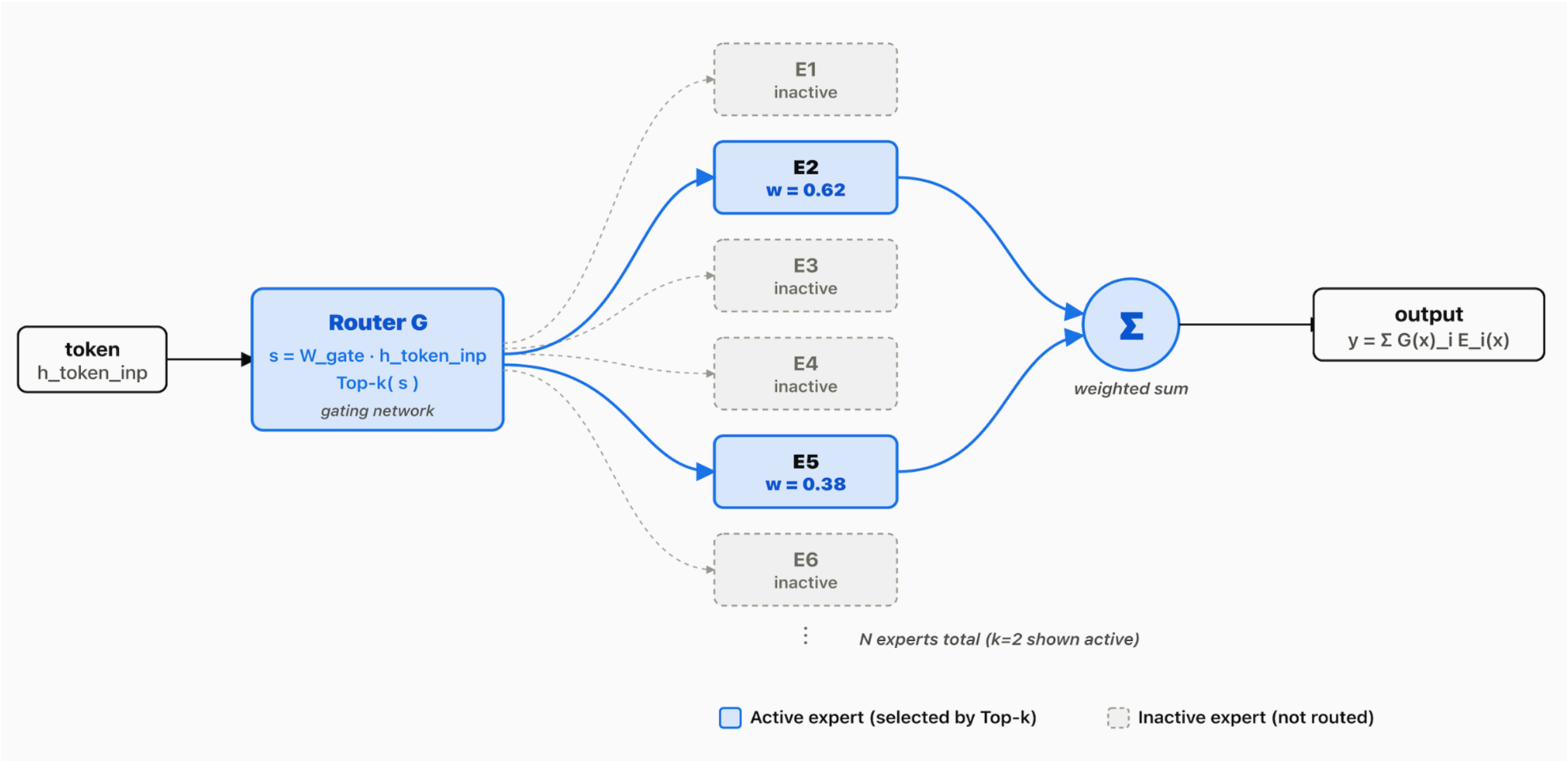

**Figure 1:** Sparse Mixture-of-Experts layer with top-k routing over N expert FFN subnetworks.

Architecturally, standard MoE models replace the static, uniform FFN layers found in dense Transformers with a set of parallel, specialized expert subnetworks E1, E2, …, EN and a parameterized routing (or gating) network G. In a Mixture-of-Experts (MoE) architecture as shown in Figure 1, a router computes a sparse gating distribution g(x) to direct inputs to a subset of k active experts out of N total experts, limiting active compute [2]. While standard implementations use small top-k expert selection such as k=1 [3] or k=2 [24], fine-grained designs scale to larger selections like k=8 over N=64 [11].

However, existing tools provide only partial support for interactive, cross-architecture, token-level visualization of MoE routing pathways. For a model with L layers and N experts per layer, a single token can follow one of many possible routing paths through the model. In fine-grained routing configurations such as OLMoE (16 layers × 64 experts = 1024 total nodes) [11] with top-8 routing, the number of possible routing trajectories grows combinatorially.

Tracing how multiple tokens interact, diverge, and trigger specialized memory blocks across a full model produces a volume of routing data that is both slow to render at full resolution and difficult for human experts to interpret directly from raw mathematical tensors.

### *1.1 Design Principles*

Our visualization framework is governed by five core design principles:

***Empirical Fidelity:*** All visualized data points are extracted directly from real model forward passes. They are not generated using synthesized, simulated, or purely illustrative representations.

***Architectural Agnosticism:*** To prevent hard-coded assumptions about expert counts or routing parameters, the visualization layouts dynamically adapt to diverse MoE topologies. The framework is designed to adapt across distinct routing granularities, layer counts, and shared-expert designs, as demonstrated on three architecturally different models. The interface dynamically adjusts to the contrasting architectures of our validation models (see specifications in Table 2).

***Transparent Resolution Scaling:*** To keep large tensor visualizations readable, MoEXViz renders high-dimensional matrices as compact heatmaps. When downsampling is required, block-averaging is used so that each visible heatmap cell summarizes a block of original tensor entries. The original tensor shape is displayed alongside each heatmap to indicate the source matrix being represented.

***Diagrammatic Traceability:*** The MoEXViz interface maintains a direct, interactive link between visual diagrams and their underlying mathematics. Selecting an interactive diagram component exposes the algebraic equation and the concrete, real-valued operands that computed it.

***Semantic Color Mapping:*** To minimize cognitive load, color is reserved strictly to distinguish between different computational stages. Quantitative metrics (such as those in the attention-map and activation-rate heatmaps) are encoded using separate, dedicated sequential or diverging colors to prevent visual conflation.

# 2 Related Work

Existing interactive visualizations for language models broadly fall into two categories: MoE-specific diagnostics and dense Transformer visualizers.

***MoE-specific Diagnostics:*** The EMO Interactive Visualizer [16] maps semantic domains to expert modules for a single production network; the theoretical analysis by Chen et al. [14] visualizes 2D t-SNE [26] routing boundaries; MoE-Visualization-GPT20B [18] records routing-choice shifts across prompt variations; and MixtureKit [19] provides a Streamlit training-time dashboard. However, these tools are either hard-coded to a single model's static traces, restricted to offline static plots, or designed for active training-run monitoring.

***Dense Transformer Visualizers:*** Tools such as BertViz [20], LLM Visualization [21], Transformer Explainer [22], AnimatedLLM [23], and Build An LLM [25] excel at showcasing attention mechanisms and matrix multiplications. Despite their utility, none of these tools support sparse architectures or input-conditional routing; they model dense layer operations only.

MoEXViz bridges these gaps by combining token-level auditability, offline extraction, multi-lens visualization, and cross-architecture support in a single interactive platform.

## *2.1 Evolution of Visualizations*

Table 1 compares various visualization tools and analytical approaches for neural network models across key capabilities like Mixture-of-Experts (MoE) support, token-level auditing, forward pass, cross-architecture support, and visual interface. Earlier approaches such as theoretical specialization papers, post-hoc diagnostics, and general training frameworks rely heavily on static aggregates or lack cross-architecture flexibility. In contrast, MoEXViz offers low-latency interactive debugging across diverse MoE models. The limitations of existing visualization and interpretability tools and how MoEXViz addresses them are summarized below.

**Table 1:** Comparative Landscape of LLM Interpretability and Analysis Tools

| Tool | Scope | MoE Support | Token Audit | Forward Pass Trace | Cross-Architecture Support | Visual Interface |
|---|---|---|---|---|---|---|
| **Specialization Papers [15], [17]** | Theoretical formulation & offline analysis | Yes | No | No | No, except [17] | Static paper tables & PDF figures |
| **MoE Diagnostics [14], [16], [18]** | Post-hoc single-experiment evaluation | Yes | Limited, [18] supports prompt variation; [14] and [16] do not | No, precomputed traces only | No, hard-coded to target model | 2D t-SNE [26] coordinates, post-hoc heatmaps |
| **MixtureKit [19]** | MoE composer + post-hoc visualizer | Yes | Yes, per-token, per-layer expert color-coding | Yes | No, limited to its own composition framework | Per-token expert color map per-layer weight tables |
| **Dense Visualizers [20], [21], [22], [23], [25]** | Transformer architecture | No | Yes, dense attention layers | Yes, in-browser inference | N/A, does not model MoEs | 3D tensor shapes, attention head maps |
| **MoEXViz (Ours)** | Interactive routing debugging & auditing | Yes | Yes, click-to-token samples with overflow count | Yes, offline forward pass + interactive replay | Yes, models OLMoE, JetMoE, DeepSeek-MoE | Expert UMAP, activation grids, trends, and rankings |

### 2.2 Limitations of Existing Visualization Tools

To contextualize the contributions of MoEXViz, we evaluate the landscape of current interpretability tools across four key dimensions, highlighting the critical gaps left by both standard dense-network profilers and existing Mixture-of-Experts (MoE) diagnostics.

***Token-Level Audit & Context:*** *Standard* hardware profilers focus on systems-level metrics such as FLOPs, latency, and memory usage, while dense Transformer visualizers provide token-level views of attention and matrix operations. However, neither category directly supports sparse MoE routing. Existing MoE-specific diagnostics remain restricted in other ways — EMO, for instance, is locked to a single fixed, precomputed clustering of a static document corpus, evaluated on one architecture.

MoEXViz bridges this gap by providing token-level traceability spanning six semantic domains, replicated identically across three architecturally distinct models, with every reported metric traceable back to the tokens that produced it. In the Domain Specialization Analysis view, MoEXViz displays a representative sample of the 40 tokens with the highest routing scores for each metric, while separately reporting the true total count.

***Routing Architecture Support:*** Traditional profiling tools designed exclusively for monolithic, dense networks are unable to model sparse branching or gating dynamics. Some MoE-specific diagnostics that exist are locked to a single model instance, while others, like MixtureKit, can compose across multiple base models but only within their own composition framework. MoEXViz introduces a cross-architecture approach built from the ground up to evaluate and compare routing behaviors across diverse, heterogeneous MoE architectures, demonstrated against three distinct, pretrained models detailed in Table 2.

***Analytical Lenses:*** Dense Transformer visualizers provide rich views of dense-model computation, but they do not represent sparse MoE expert routing, per-token expert selection, gating weights, or cross-domain expert specialization. Existing MoE diagnostics similarly lack a synchronized, domain-level view: even tools with multiple panels, such as MixtureKit's per-token color map and weight table, stop at a single prompt's routing trace, with no cross-prompt, cross-domain lens. MoEXViz provides a unified multi-lens interface, delivering synchronized, complementary views that simultaneously expose spatial similarity, per-expert grids, depth trends, and domain rankings.

***Temporal & Depth Trajectory:***

Mapping the dynamic path of a token through a sparse MoE network remains challenging. Dense Transformer visualizers do not represent sparse expert-routing pathways, while existing MoE-specific tools either rely on static post-hoc heatmaps and t-SNE grids [26] or expose routing behavior one layer at a time rather than as a connected trajectory. As a result, the progression of token routing across depth is often collapsed or fragmented. MoEXViz addresses this by animating layer-by-layer routing replay, narrating how a token's or the full sequence's routing decisions and residual state connect from one layer to the next.

## 3 MoEXViz Overview

MoEXViz is a visualization platform designed for both low-granularity and fine-grained MoE architectures. The platform provides the following features, each addressing a specific limitation identified.

***Auditable Visualizations:*** Existing approaches [14], [15], [16], [17] assess model specialization through aggregate, post-hoc statistics but offer no mechanism for independent verification. We address this limitation by coupling statistical aggregation with granular auditability at two levels.

In the Domain Specialization Analysis tab under the "Expert activation rates per layer" Histogram view, each (domain, layer, expert) cell is backed by an explicit index from the reported statistic to the underlying tokens. Clicking a histogram bar reveals the derived metrics: expert activation rate, normalized rate, average softmax probability, and specialization score. The popup also displays a sample of the actual tokens routed to produce these metrics, with the remaining count displayed when the set exceeds the 40-token display cap.

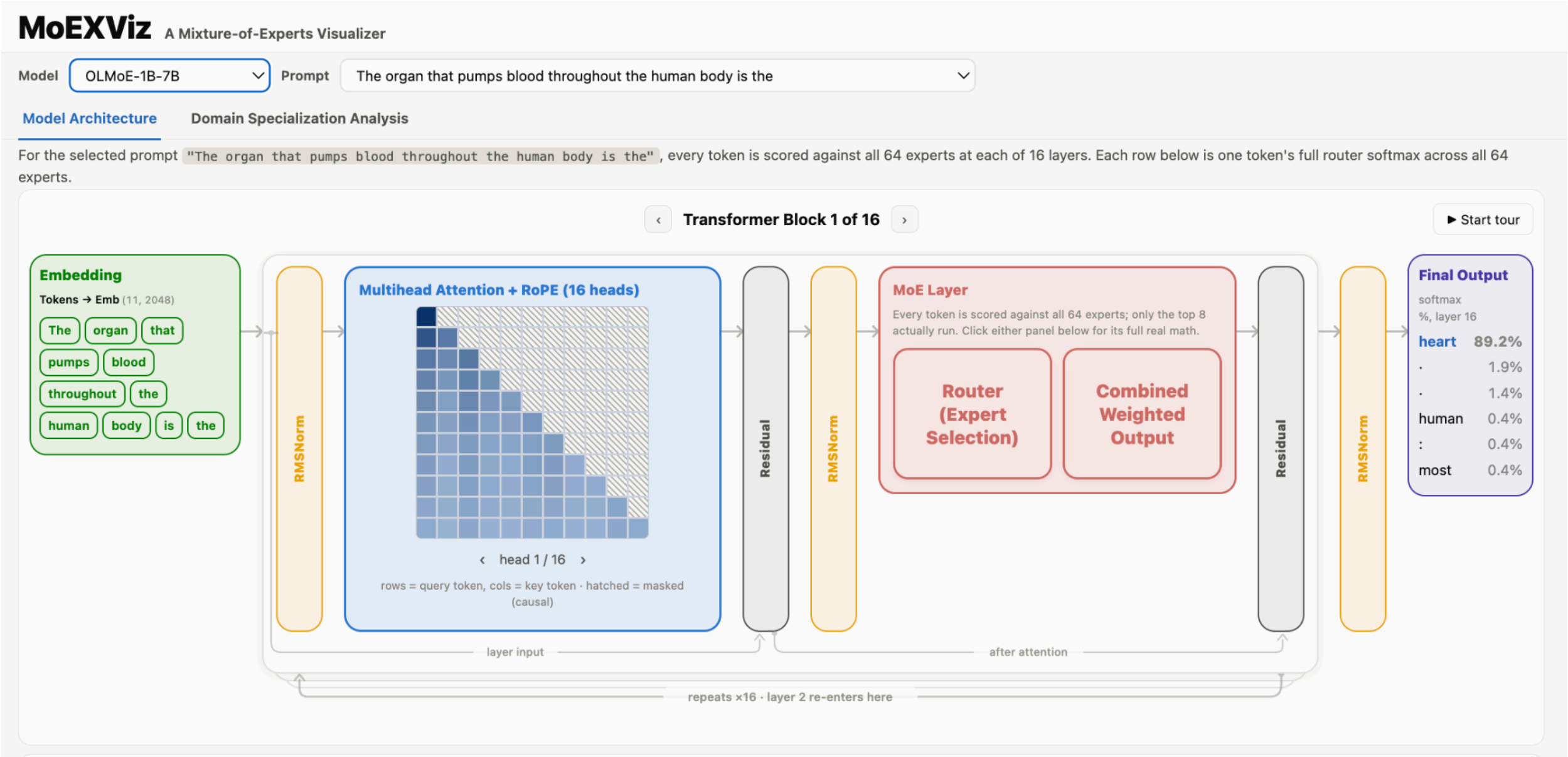

**Figure 2:** Model Architecture view for OLMoE-1B-7B. The interface shows the token flow through embedding, attention, RoPE, MoE routing, weighted expert aggregation, and final output prediction.

In the Model Architecture view (Figure 2), this same auditability extends to an individual token's full path through the network: stepping layer-by-layer reveals that token's real top-k expert selections and softmax weights at each layer, down to the model's actual next-token prediction at the output. Together, these let a user spot-check both an aggregate specialization claim and a single token's routing trajectory against raw model outputs.

***Built-in Cross-Architecture Comparison:*** Our interface natively supports a diverse suite of Mixture-of-Experts (MoE) architectures, including OLMoE-1B-7B [11], JetMoE-8B [12], and DeepSeek-MoE-16B [13]. By providing unified visualizations across these models, MoEXViz enables comparison of specialization patterns across architecturally distinct MoE systems.

***Clear Visualization of Expert Selection Logic and Architecture*:** In the Model Architecture view shown in Figure 2, users can step through layers to inspect both a single token's routing decision and its relationship to the full prompt.

Clicking the Router Block opens the Single Token Routing sub-tab, where selecting a token reveals the underlying arithmetic for its selected top-k experts. Switching to the All Tokens Routing sub-tab renders every input token's expert-activation profile for the current layer. This allows users to quickly compare whether different tokens converge on the same experts or diverge across layers.

For each token, the Combined Weighted Output view shows the actual weighted-sum equation over the selected top-k expert outputs, resulting in the final MoE output vector.

***Sparse Routing in Dense Context:*** The Model Architecture tab in MoEXViz, shown in Figure 2, connects router behavior back to the real attention, RoPE, RMSNorm, and SwiGLU computations upstream of it. This is a capability that general Transformer visualization tools, such as Transformer Explainer, LLM Visualization, and BertViz, do not provide because they render dense attention rather than sparse expert routing.

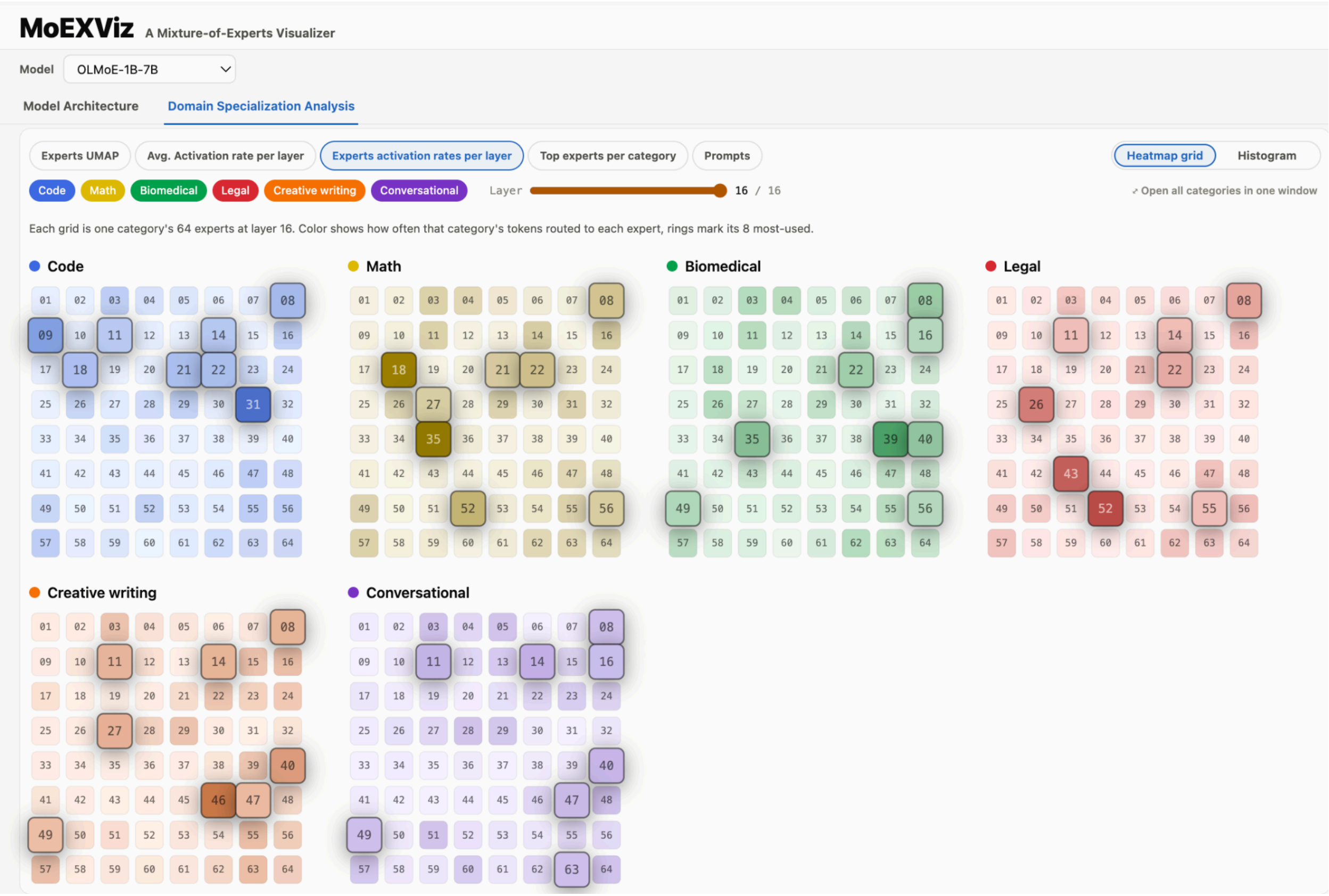

**Figure 3:** Domain Specialization Analysis for OLMoE-1B-7B at layer 16. Each 8×8 grid shows expert activation rates for one domain. Darker cells indicate higher activation rates, and outlined cells mark the eight most-used experts for that domain.

***Domain-Level Specialization Analysis:*** MoEXViz also explores whether routing varies systematically by input domain, which is central to whether specialization is a meaningful, semantic property or just a load-balancing artifact. The Domain Specialization Analysis tab (Figure 3) runs the same six-domain probe set (Code, Math, Biomedical, Legal, Creative Writing, Conversational) across all three models, exposed through five complementary views: an Expert UMAP, Avg activation rate per layer, Expert activation rates per layer, Top experts per category, and the source prompts. These let a user judge, model by model, whether a domain's tokens concentrate on a small expert subset or spread diffusely, and compare that pattern directly across architectures since the same protocol runs on all three.

At the final layer, biomedical and legal prompts show divergence in their top-k activated experts across all three models. Let topk(d,l) be the top-k experts for domain d at layer l. Divergence is then defined as:

$$divergence(d1,\ d2,\ l) = k - |topk(d1,\ l) \cap topk(d2,\ l)|$$

For OLMoE-1B-7B, 6 of 8 experts differ, 1 of 2 for JetMoE-8B, and 2 of 6 for DeepSeek-MoE-16B (Figure 3 shows this for OLMoE).

With five prompts per domain in the current dataset, this is a consistent example, though not yet a statistically established finding. Paired with the click-to-token auditability described above, this turns specialization from an aggregate claim into an inspectable one, traceable down to the tokens behind it.

## 3.1 Model Choice for Visualization

These models represent contrasting points in the MoE design space and were chosen because their routing designs differ sharply. As detailed in Table 2, OLMoE-1B-7B provides a testbed for fine-grained, high-count routing; JetMoE-8B evaluates routing applied to both attention and FFN blocks; and DeepSeek-MoE-16B tests the visualization's ability to isolate always-on shared experts from standard routed experts. Selecting these diverse topologies is intended to exercise the visual abstractions across a broader range of MoE routing patterns.

In particular, OLMoE-1B-7B represents the fine-grained extreme, with 64 small experts per layer and top-8 routing; JetMoE-8B represents the low-granularity structure with only 8 larger experts per layer and top-2 routing, and also applies sparse routing to its attention block, not just its FFN layer; DeepSeek-MoE-16B adds a third routing pattern: alongside 64 routed experts with top-6 selection, it includes 2 shared experts that are always active for every token.

# 4 System Design

### *4.1 Pipeline and Data Extraction Architecture*

To provide deep interpretability without requiring high-performance computing resources or a live GPU connection at visualization time, MoEXViz implements a three-stage pipeline: (i) Offline Extraction, (ii) JSON Serialization, and (iii) Interactive Client-Side Playback. In this workflow, model inference is performed during the offline extraction stage; the browser does not execute the model or require GPU access, but instead parses and visualizes precomputed routing traces stored in JSON. This decoupling allows MoEXViz to run entirely in a browser on consumer hardware once the one-time offline extraction step has produced its JSON payload. Figure 4 illustrates this pipeline end-to-end.

### *4.2 Experimental Testbeds*

We evaluate MoEXViz across three open-weight Mixture-of-Experts models: OLMoE-1B-7B [11], JetMoE-8B [12], and DeepSeek-MoE-16B [13].

To accommodate the topological differences, each model's rendering page embeds that model's configuration metadata, layer count, expert count, and top-k routing width directly in its data object at export time. At render time, this embedded configuration drives the CSS grid dimensions and the hand-written SVG coordinate functions that lay out the router diagram, top-k highlights, and domain specialization analysis grids; the same rendering code determines their grid sizes, normalization factors, and coordinate mappings from the expert count and top-k value read from that metadata.

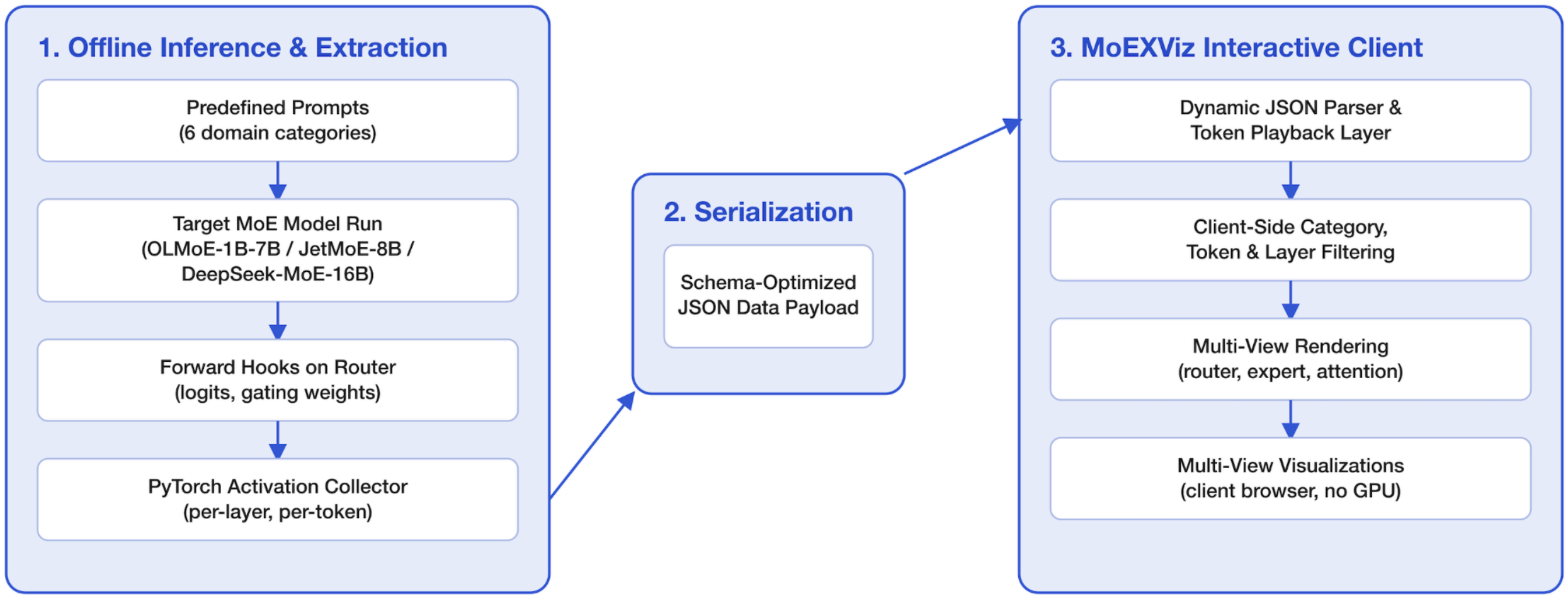

**Figure 4: MoEXViz three-stage pipeline:** Offline inference collects router activations, serialization stores compact JSON traces, and the browser client renders interactive visualizations without requiring live GPU access.

**Table 2**: Architectural specifications of the selected MoE models.

| Architectural Dimension | OLMoE-1B-7B [11] | JetMoE-8B [12] | DeepSeek-MoE-16B [13] |
|---|---|---|---|
| Total Parameters | ~6.9 Billion | ~8.5 Billion | ~16.4 Billion |
| Active Parameters (Inference) | ~1.3 Billion | ~2.3 Billion | ~2.8 Billion |
| MoE Configuration | 16 MoE layers | 24 MoE layers | 1 dense FFN layer + 27 MoE layers |
| Experts (MoE Layers) | 64 | 8 | 64 |
| Routing (top-k) | top-8 | top-2 | top-6 |
| Shared Experts | None | None | 2 |
| Sparse Routed Blocks | FFN | Attention & FFN | FFN |
| Attention Head Configuration | 16 heads | 32 query heads (GQA, 16 shared K/V heads); attention routed over top-2 of 8 attention experts. | 16 heads |

### *4.3 Activation Capture and Forward Hooks*

To extract routing behavior, we run each target model on prompts spanning six semantic domains: Code, Biomedical, Math, Legal, Creative Writing, and Conversational. We use two prompt sets: a 12-prompt set for detailed architecture playback and a 30-prompt set for domain-level specialization analysis.

For an input sequence of length $T$, PyTorch forward hooks are registered to the router modules of each layer. During the forward pass, these hooks intercept and log routing information for every token t at layer l. The primary extracted variables are:

***Expert Indices:*** A subset containing the unique identifiers of the top-k selected experts.

***Gating Weights:*** The routing weights assigned to the selected top-k experts and used in the model's weighted expert aggregation.

**Dense Router Scores:** The full pre-top-k softmax distribution over all N experts, used for domain-level metrics such as average softmax probability.

For the Model Architecture view for JetMoE-8B, which applies sparse routing to both attention and feed-forward blocks, we treat the attention router and FFN router as separate routed components and serialize their routing traces independently. This allows the visualization to distinguish whether a token's sparse path changes in the attention block, the FFN block, or both.

### *4.4 Metrics for Domain Specialization Analysis*

In the current dataset, each of the six domains contains five prompts.

**Expert activation rate:** For a domain d, let T(d) denote the total number of tokens across that domain's pooled passages. Let $hits(d,\ l,\ e)$ be the number of domain $d$'s tokens routed to expert e at layer l; then

$$activation_rate(d,\ l,\ e) \ = \ hits(d,\ l,\ e) \div T(d)$$

Used in: the click-to-token drill-down popup within the Expert activation rates per layer tab's Histogram view.

***Normalized rate:*** Expert activation rate divided by k, so summing across all $N$ experts at a fixed (domain, layer) yields 100%.

$$normalized_rate(d,\ l,\ e) \ = \ [activation_rate(d,\ l,\ e) \div k] \times 100\%$$

Used in: the click-to-token drill-down popup within the Expert activation rates per layer tab's Histogram view.

***Average softmax probability:*** the mean softmax probability that domain d's tokens assign to expert $e$ at layer l, whether or not the expert was selected.

*avg_softmax_prob(d, l, e)* = mean over domain d's *T(d)* tokens of each token's full softmax probability for expert e at layer l.

Used in: the click-to-token drill-down popup within the Expert activation rates per layer tab's Histogram view.

***Specialization score***: the $log_2$ fold-change of a domain's rate against the six-domain mean at that (layer, expert), with smoothing constant $\varepsilon$ = *0.0001*. Here, *avg_rate(l, e)* is equal to the average of *activation_rate(d', l, e)* across all six domains *d'*.

$$spec_score(d, l, e) = log_2 [ (activation_rate(d, l, e) + \varepsilon) \div (avg_rate(l, e) + \varepsilon)]$$

Used in: the click-to-token drill-down popup within the Expert activation rates per layer tab's Histogram view.

***Average activation rate per layer:*** the activation rates of a domain's k most-used experts at each layer, averaged into a single scalar(drives the per-layer trend line).

*Average activation rate per layer(d, l)* = average of $activation_rate(d, l, e)$ over the top-k experts e at layer *l*.

Used in: the Avg activation rate per layer tab.

***Top experts per category:*** ranks layer-expert pairs by raw activation rate. Unlike the specialization score, this metric indicates how often a domain uses an expert, not whether it uses that expert more than other domains.

$top_experts_per_category(d)$ = the (layer, expert) pairs with the highest *activation_rate(d, l, e)*, sorted descending, top 8 kept.

Used in: the Top experts per category tab.

***Expert UMAP:*** projects each (layer, expert) pair's activation profile across all six domains:

$$v(l, e) = ( activation_rate(code, l, e), activation_rate(math, l, e), activation_rate(biomedical, l, e), activation_rate(legal, l, e), activation_rate(creative\ writing, l, e), activation_rate(conversational, l, e))$$

$v(l, e)$ is a 6-dimensional vector, projected down to 2D via UMAP [27] (cosine similarity metric, *n_neighbors = 15*, *min_dist = 0.1*). This excludes pairs that never activate on any domain (all-zero vectors). Each point is additionally assigned a fixed color based on its dominant domain, defined as the domain with the highest value in its six-dimensional activation vector.

Used in: the Expert UMAP tab.

### *4.5 Lightweight JSON Serialization Schema*

Collecting raw tensor activations across prompts and layers produces gigabytes of data, far more than a browser can efficiently fetch. We split serialization by need: each token's router output is kept dense (softmax over all N experts), so non-selected experts still show a real routing score; FFN weight matrices, by contrast, are serialized only for the sparse set of (layer, expert) pairs activated by at least one token, omitting experts that are never used in the analyzed prompt set. For models with shared experts, such as DeepSeek-MoE-16B, shared-expert weights and outputs are serialized unconditionally, since these experts are active for every token and have no unused case to omit. All retained matrices (router, expert, and attention weights) are then downsampled by block-averaging before being written to JSON.

The result is a single schema-optimized JSON object per prompt, each tagged with a domain label and containing the tokenized prompt. For each layer, the top-k expert indices, their gating weights, and the downsampled weight/activation grids needed to render that layer's details are all stored. Thus the browser can retrieve any serialized layer-token-expert triple associated with the analyzed prompt set without re-running the model or re-deriving routing values at render time.

### *4.6 Dynamic Derivations*

Our approach avoids hard-coded layout assumptions by re-deriving every number and view from the selected model's own configuration whenever the user switches the model selector.

When a user loads the identical prompt (*"The most popular programming language for data science is"*), the interface dynamically adapts its math and layouts depending on the active model selector. For OLMoE-1B-7B, it structures an evaluation grid spanning 64 experts across 16 layers; for JetMoE-8B, it dynamically adds a second router visualization to handle parallel attention and FFN experts; and for DeepSeek-MoE-16B, it dynamically recalculates coordinates to isolate the always-on shared experts across 27 MoE layers.

The same pattern holds in the Router view shown in Figure 5. A hub-and-spoke diagram connects a central Router node to every expert for the current token. For DeepSeek-MoE-16B, the visualization highlights and weights the token's selected top-6 routed experts.

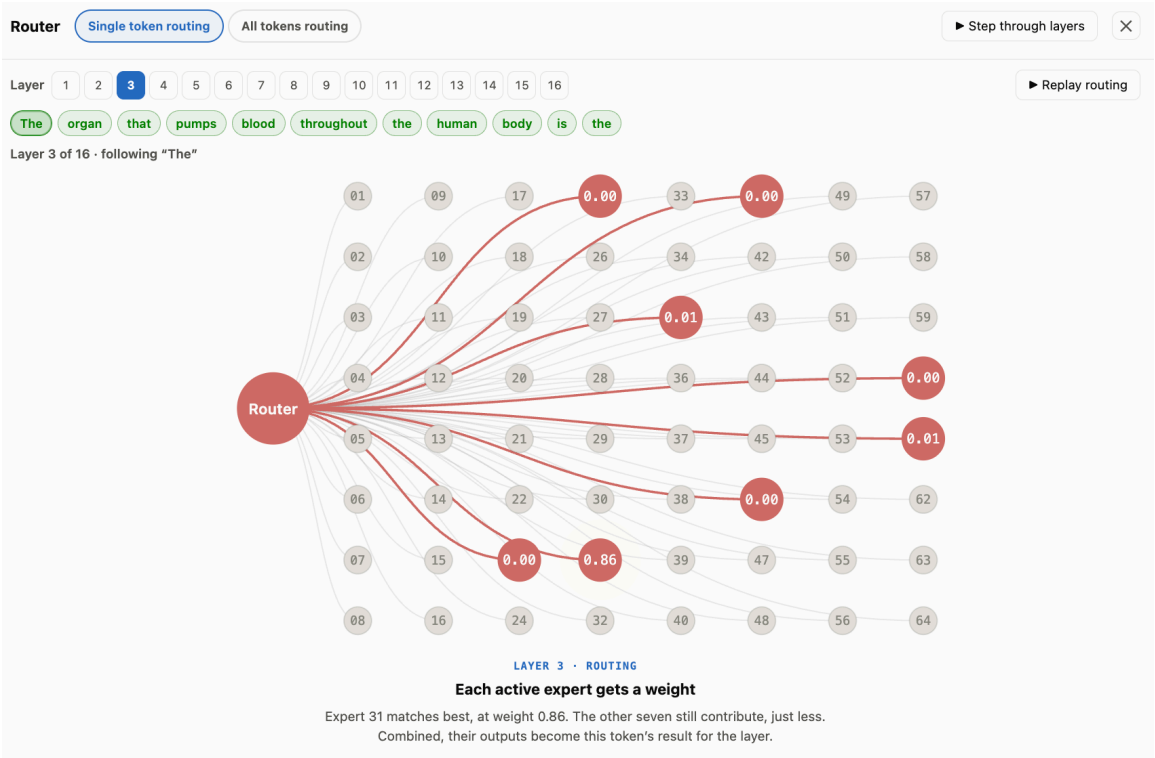

**Figure 5:** DeepSeek-MoE-16B router visualization. A central router node connects to 64 routed experts arranged in an 8×8 grid, with the top-6 selected experts highlighted for the current token.

The Parameter-Count Panel derives its per-expert and per-layer arithmetic dynamically, computing parameter allocations on the fly by querying the loaded model's tensor dimensions and configuration keys (hidden size, intermediate size, expert count, top-k, shared-expert count). For OLMoE and DeepSeek-MoE-16B, the panel's summary sentences additionally report each model's published total and active-parameter counts for user context; these two figures are not re-derived from the FFN-only tensor shapes captured in the trace. JetMoE-8B is the exception: its total and active-parameter summary figures are computed live from the same trace-derived tensor dimensions as the per-expert/per-layer arithmetic above.

For example, when inspecting DeepSeek-MoE-16B [13], the panel programmatically computes and outputs:

***Parameters per routed expert*** *(representing the gate, up, and down projection matrices of the SwiGLU FFN layer)**:***
Params_expert = 3 × (2048 × 1408) = 8,650,752

***Parameters per MoE layer*** *(accounting for both routed and shared experts):*
Params_layer = (64 routed + 2 shared) × 8,650,752 = 570,949,632

These calculations refer specifically to MoE parameters: routed feed-forward experts for OLMoE-1B-7B and DeepSeek-MoE-16B, plus routed attention-expert and shared key/value projections for JetMoE-8B's dual-router design. They exclude non-expert parameters such as embeddings, non-routed attention projections, normalization layers, and the output head unless otherwise stated.

# 5 Limitations

***Lossy downsampling:*** Large matrices are block-averaged before serialization, so displayed grid cells are regional means, not individual weight or activation values, and the tool supports inspecting patterns, not exact per-parameter values.

***Fixed prompt set, no live inference*****:** The Model Architecture tab (Figure 2) uses 12 precomputed prompts, with two prompts per domain, while the Domain Specialization Analysis tab uses a larger probe set of 30 prompts, with five prompts per domain.

***Small domain sample size:*** The domain-specialization views are based on five prompts per domain (prompts used are displayed in the Prompts tab). As a result, observed expert-use patterns should be interpreted as examples from the analyzed prompt set and not definitive evidence of stable domain specialization across the model.

***Limited user evaluation:*** MoEXViz currently provides a system-level demonstration rather than a full human-subject evaluation.

***Limited architecture coverage:*** Generality claims rest on three models (OLMoE-1B-7B, JetMoE-8B, DeepSeek-MoE-16B); other MoE variants (e.g., expert-choice or soft-routing schemes) remain untested, and the schema's assumptions may not transfer directly to them.

# 6 Conclusion and Future Work

This work presents an open-source, interactive visual analysis tool that addresses a critical gap in Mixture-of-Experts (MoE) analysis by integrating per-token audit trails, linked cross-view contexts, and upstream routing mechanics into a single, cohesive interface. Practical auditability is achieved through a robust three-stage pipeline in which every displayed routing metric traces back to a real model forward pass, while large matrix views are explicitly shown through documented block-averaged downsampling. To demonstrate generality, this design was applied to three architecturally diverse models: OLMoE-1B-7B, JetMoE-8B, and DeepSeek-MoE-16B.

As future work, we intend to conduct empirical user studies to substantiate the tool's practical debugging utility and also support additional MoE architectures, particularly those with routing mechanisms that differ substantially from the models studied here.